\documentclass[11pt,a4paper]{article}
\usepackage[hyperref]{acl2019}
\usepackage{times}
\usepackage[capitalize,noabbrev]{cleveref}
\usepackage{latexsym}
\usepackage{rotating}
\usepackage{graphicx}
\usepackage{multirow}
\usepackage{color}
\usepackage{url}
\usepackage{tikz}
\usepackage[utf8]{inputenc}
\usepackage[T1]{fontenc}
\usepackage{booktabs}
\usepackage{tweaklist2012}

\aclfinalcopy 
\def\aclpaperid{***} 

\def\perscite#1{\citet{#1}} 
\def\parcite#1{\citep{#1}} 

\newlength{\oboshortcutssaveparindent}
\def\squeezelist{
\renewcommand{\itemhook}{
  \setlength{\itemsep}{0pt}
  \setlength{\parsep}{0em}
  \setlength{\topsep}{0pt}
  \setlength{\partopsep}{0pt}
  \setlength{\leftmargin}{1em}
  \setlength{\labelwidth}{1em}
  \setlength{\labelsep}{0.5em}
}
}

\newlength{\oboshortcutssaveparindentenum}
\def\squeezeenum{
\renewcommand{\enumhook}{
  \setlength{\itemsep}{0pt}
  \setlength{\parsep}{0em}
  \setlength{\topsep}{0pt}
  \setlength{\partopsep}{0pt}
  \setlength{\leftmargin}{1em}
  \setlength{\labelwidth}{1em}
  \setlength{\labelsep}{0.5em}
}
}

\def\system#1{#1} 
\def\onlB{\system{online-B}}
\def\docTr{\system{C-DocTrafo-T2T-2019}} 
\def\onlA{\system{online-A}}
\def\onlG{\system{online-G}}
\def\onlX{\system{online-X}}
\def\onlY{\system{online-Y}}
\def\TrNineteen{\system{C-Trafo-T2T-2019}}
\def\TrEighteen{\system{C-Trafo-T2T-2018}}
\def\TrMarian{\system{C-DocTrafo-Marian}}
\def\uedin{\system{uedin}}
\def\tartu{\system{TartuNLP-c}}

\def\to{$\rightarrow$}

\def\Tref#1{Table~\ref{#1}}

\title{SAO WMT19 Test Suite: Machine Translation of Audit Reports}

\author{Tereza Vojt\v{e}chov\'{a}*
        \qquad Michal Nov\'{a}k*
        \qquad Milo\v{s} Klou\v{c}ek*${}^{\dag}$
        \qquad Ond{\v{r}}ej Bojar*
		\\ \\
       {}* Charles University, Faculty of Mathematics and Physics \\
        Institute of Formal and Applied Linguistics \\
        Malostransk{\'{e}} n{\'{a}}m{\v{e}}st{\'{\i}} 25, 118 00 Prague, Czech Republic \\
         \\
         ${}^{\dag}$ Supreme Audit Office of the Czech Republic \\
         Jankovcova 1518/2, 170 04 Prague, Czech Republic \\
         {\tt \{vojtechova, mnovak, kloucek, bojar\}@ufal.mff.cuni.cz}}

\date{}

\begin{document}
\squeezelist 
\squeezeenum{}
\maketitle
\begin{abstract}
This paper describes a machine translation test set of documents from the auditing domain and
its use as one of the ``test suites'' in the WMT19 News Translation Task for
translation directions involving Czech, English and German.

Our evaluation suggests that current MT systems optimized for the general news
domain can perform quite well even in the particular domain of audit reports.
The detailed manual evaluation 
however
indicates that deep factual knowledge of the domain is necessary. For
the naked eye of a non-expert, translations by many systems seem almost perfect
and automatic MT evaluation with one reference is practically useless for
considering these details.

Furthermore, we show on a sample document from the domain of agreements that
even the best systems completely fail in preserving the semantics of the
agreement, namely the identity of the parties.
\end{abstract}

\section{Introduction}

Domain mismatch is often the main sources of machine translation errors. At the
same time, it has been suggested in the speech recognition area that models
trained on extremely large data can perform well across domains, i.e. without
any particular domain adaptation \parcite{domain-invariant-asr:2018}.

We believe that for some of the language pairs annually tested in the WMT shared
translation task, the best machine translation systems may have grown to sizes
where the domain dependence may be less critical. At the same time, we know that
most of current MT systems still operate at the level of individual sentences
and therefore have no control over document-level coherence e.g. in terms of
lexical choice.

To investigate the two questions, domain independence and document-level
coherence, we cleaned and prepared a dedicated set of documents from the
auditing domain and submitted it as one of the ``test suites'' to this year's
WMT News Translation Task. The collection is called ``SAO WMT19 Test Suite''
after the Supreme Audit Office of the Czech Republic (SAO)
who provided the
original audit reports created in cooperation with other national supreme audit
institutions (SAIs).\footnote{We adhere to
the convention that ``SAO'' refers solely to the Supreme Audit Office of the
Czech Republic. For other supreme audit institutions, we use the acronym SAI.}


This paper is organized as follows: In \cref{composition} we describe the source
and our processing of the test documents. \cref{auto} provides automatic scores
of WMT19 MT systems on the test suite and \cref{sec:evaluation} presents the
manual evaluation.
One more document type, namely a sublease agreement, was evaluated separately,
see \cref{sec:sml}. We release the test suite for public use, see
\cref{availability}, and
we conclude in \cref{conclusion}.




\section{Composition of SAO Test Suite}
\label{composition}

The SAO Test Suite consists of 10 multi-language audit reports issued by
the SAO. The reports describe investigations carried out jointly by SAO and
one or more other national auditing institutions between the years 2004 and
2015. The reports were published in multiple language versions or as
multilingual documents.
They were created jointly by the co-operating SAIs in English and later on, they were translated by translation agencies and finally corrected by the authorized auditors from the respective countries.
The end
effect of this
careful procedure is that 
from time to time,
the different language versions slightly depart in the exact wording, including
minor shifts of the conveyed meanings.

All the reports come in 3 different languages.
All of them include Czech and English, the third used language differs. See
\Tref{tab:languages} for a summary.

\begin{table}[t]
\centering
\begin{tabular}{lc}
{Language} & {Count} \\ 
\hline
\bf Czech &  10      \\
\bf English & 10 \\
Slovak & 5 \\
\bf German & 4 \\
Polish & 1 \\
\hline
Total documents & 30 \\
\end{tabular}
\caption{Number of languages in SAO Test Suite. Languages in bold were used in
WMT19 Shared Translation Task.}
\label{tab:languages}
\end{table}


\subsection{Creation of the SAO Test Suite}

The audit reports were collected primarily from the website of SAO.
It is important to note that while being publicly available, these documents did
not make it to any of WMT19 constrained training data,
probably
because the texts appear on the web only in the form of PDFs.
We double-checked that there is no overlap by
searching the data for exact and near sentence matches. Very short segments like
generic titles or section numbers were naturally present in the training data
but we did not find any longer sentences, let alone more sentences from a test
document.

First, we converted the documents from the PDF format to plain text. We note that some of the documents were bitmap PDFs (scans) and we had to use OCR to
obtain the text. This was particularly tedious for multi-language documents
with texts side by side in two or three columns.

The rest of the processing was applied only to Czech, English and German
versions of the documents, because other languages were not considered in WMT19
News Translation Task.

The plain text versions were
automatically segmented into sentences using the trainable
tokenizer TrTok by \perscite{marsik:bojar:pbml:2012}.
We then automatically aligned sentences in English and Czech versions using
hunalign \parcite{varga:hunalign:2005} and manually revised this alignment.

During the manual revision of sentence alignments, we removed footnotes, tables
and graph captions, as well as occasional paragraphs not present in one of the
languages. Sometimes, sentence segmentation had to be fixed as well.

In the final stage, we added the German side to the already sentence-aligned
English-Czech files, creating a tri-parallel test set. In some cases, the
segmentation into sentences was not exactly parallel and we had to break
primarily the German sentences into clauses, or introduce blank segments in some
of the files to allow for a better match.
Once or twice even the order of the clauses in German was swapped compared to
the aligned Czech and English.

\subsection{SAO Test Suite in WMT19 Shared Task}

We submitted our files as a ``test suite'' complementing
the WMT19 News Translation Task. This means that all primary MT systems
participating in the News Translation Task also translated our files.

The English\to{}Czech and German$\leftrightarrow$English systems were
\emph{supervised}, i.e. trained on genuine parallel texts (and target-side
monolingual data). The Czech$\leftrightarrow$German research systems were
\emph{unsupervised}, i.e. trained only on monolingual source and target texts,
optionally using a small parallel development set of a few thousand sentence
pairs. Our evaluation also includes several anonymized online systems
(``online-\dots'') the internals of which are not known. These online systems
could in principle include our test suite as part of their training data.

The number of evaluated documents and MT systems for each examined language pair
is in \cref{nums}.

\begin{table}[t]
\begin{center}
\begin{tabular}{lcc}
Language Pair&Documents&MT Systems\\
\hline
en-cs        &11       &11\\
en-de        &4        &22\\
de-en        &4        &16\\
cs-de        &4        &7\\
de-cs        &4        &11\\
\end{tabular}
\end{center}
\caption{Evaluated language pairs, documents and MT systems.}
\label{nums}
\end{table}


%
%

\section{Automatic Evaluation}
\label{auto}

For automatic evaluation, we use several of common MT evaluation metrics
\parcite{papineni:2002,popovic-2015,leusch:matr:2008,wwang:character:2016,snover:etal:2006}.
Metrics listed with the prefix ``n'' are reversed ($1-\textrm{score}$) so that
higher numbers indicate a better translation in all the figures we report.

We calculate the score for each of the documents in our test suite separately
and report the average score and the standard deviation.

The scores are detailed in Tables~\ref{auto:en-cs} to \ref{auto:de-cs}. In the
subsequent tables, we sometimes abbreviate system names for typesetting reasons.

\def\oossymb{$\wr$}
\def\oosmark{~~\llap{\oossymb{} }}

\def\scoretab#1#2#3#4{
\begin{table*}[t]
\begin{center}
\scriptsize
\begin{tabular}{lrrrrrrr}
\input{system-metric-score#4.#1.tex}
\end{tabular}
\end{center}
\caption{Automatic scores for #2. ``\oossymb'' marks scores out of sequence. #3}
\label{auto:#1}
\end{table*}
}

\begin{table*}[t]
\begin{center}
\scriptsize
\begin{tabular}{lrrrrrrr}
                          &BLEU          &chrF3                   &nCDER                   &nCharacTER              &nPER                    &nTER                    &nWER\\
\hline
CUNI-Transformer-T2T-2018 &30.21$\pm$6.22&58.49$\pm$4.14          &50.69$\pm$6.48          &50.27$\pm$9.47          &58.04$\pm$7.70          &46.20$\pm$9.04          &44.11$\pm$8.79\\
CUNI-Transformer-T2T-2019 &29.16$\pm$6.16&57.40$\pm$3.96          &49.61$\pm$6.54          &47.75$\pm$8.97          &56.76$\pm$7.93          &44.64$\pm$9.21          &42.48$\pm$8.99\\
CUNI-DocTransformer-T2T   &29.15$\pm$6.04&57.33$\pm$3.87          &49.57$\pm$6.61          &\oosmark{48.38$\pm$9.31}&56.34$\pm$7.53          &44.53$\pm$8.93          &42.45$\pm$8.78\\
uedin                     &29.15$\pm$5.94&57.31$\pm$3.84          &\oosmark{49.87$\pm$6.29}&\oosmark{48.73$\pm$8.25}&\oosmark{57.02$\pm$7.06}&\oosmark{45.53$\pm$8.53}&\oosmark{43.49$\pm$8.20}\\
online-B                  &29.14$\pm$5.57&\oosmark{57.36$\pm$3.39}&49.74$\pm$5.78          &48.44$\pm$8.23          &\oosmark{57.47$\pm$7.18}&45.46$\pm$8.22          &43.15$\pm$7.88\\
online-Y                  &28.53$\pm$5.57&57.34$\pm$3.56          &49.44$\pm$6.10          &45.00$\pm$7.78          &56.89$\pm$7.48          &45.04$\pm$8.47          &42.92$\pm$8.14\\
CUNI-DocTransformer-Marian&25.86$\pm$4.57&54.65$\pm$3.11          &46.73$\pm$5.45          &-6.50$\pm$110.46        &53.60$\pm$6.64          &41.15$\pm$7.74          &39.07$\pm$7.46\\
TartuNLP-c                &25.12$\pm$4.94&54.57$\pm$3.00          &46.21$\pm$5.80          &\oosmark{44.71$\pm$7.40}&53.02$\pm$7.92          &40.40$\pm$8.44          &38.31$\pm$8.11\\
online-A                  &24.01$\pm$5.72&53.59$\pm$3.58          &45.19$\pm$6.45          &\oosmark{44.80$\pm$8.27}&52.84$\pm$7.52          &40.27$\pm$9.03          &38.19$\pm$8.74\\
online-G                  &23.84$\pm$4.64&\oosmark{54.21$\pm$3.40}&44.78$\pm$5.79          &\oosmark{45.91$\pm$9.40}&52.83$\pm$7.02          &40.16$\pm$7.88          &38.02$\pm$7.58\\
online-X                  &19.61$\pm$3.43&50.42$\pm$2.69          &41.07$\pm$4.22          &41.39$\pm$6.72          &47.54$\pm$6.78          &34.62$\pm$6.81          &32.79$\pm$6.65\\

\end{tabular}
\end{center}
\caption{Automatic scores for English\to{}Czech. ``\oossymb'' marks scores out of sequence. }
\label{auto:en-cs}
\end{table*}

\begin{table*}[t]
\begin{center}
\scriptsize
\begin{tabular}{lrrrrrrr}
                    &BLEU          &chrF3                   &nCDER                   &nCharacTER              &nPER                    &nTER                    &nWER\\
\hline
Microsoft-sent-level&22.06$\pm$3.61&55.57$\pm$2.24          &42.62$\pm$4.68          &38.22$\pm$4.08          &44.83$\pm$5.04          &30.23$\pm$5.91          &28.37$\pm$5.89\\
Microsoft-doc-level &21.91$\pm$3.57&\oosmark{55.84$\pm$2.07}&42.52$\pm$4.50          &\oosmark{38.63$\pm$3.89}&44.18$\pm$5.42          &29.67$\pm$6.37          &27.72$\pm$6.30\\
online-B            &21.70$\pm$3.73&54.55$\pm$2.35          &41.48$\pm$4.47          &34.63$\pm$6.04          &\oosmark{46.25$\pm$5.41}&\oosmark{30.44$\pm$6.15}&\oosmark{28.61$\pm$6.17}\\
Facebook\_FAIR      &21.52$\pm$4.21&\oosmark{55.20$\pm$2.72}&\oosmark{42.24$\pm$5.17}&\oosmark{37.65$\pm$4.34}&43.49$\pm$6.16          &29.35$\pm$7.23          &27.36$\pm$7.21\\
lmu-ctx-tf-single   &21.52$\pm$3.77&54.72$\pm$2.11          &41.91$\pm$4.42          &37.50$\pm$4.86          &\oosmark{45.40$\pm$5.41}&\oosmark{30.20$\pm$6.11}&\oosmark{28.24$\pm$5.98}\\
NEU                 &21.29$\pm$3.61&54.63$\pm$1.97          &\oosmark{42.11$\pm$4.62}&\oosmark{38.36$\pm$4.45}&44.73$\pm$5.34          &30.01$\pm$6.52          &28.16$\pm$6.40\\
MSRA.MADL           &21.23$\pm$3.82&53.96$\pm$2.07          &41.20$\pm$4.65          &37.14$\pm$3.24          &44.07$\pm$5.99          &29.07$\pm$6.68          &27.29$\pm$6.59\\
Helsinki-NLP        &20.57$\pm$3.39&53.35$\pm$1.84          &41.09$\pm$4.56          &36.16$\pm$3.96          &\oosmark{44.76$\pm$5.00}&\oosmark{29.51$\pm$5.99}&\oosmark{27.65$\pm$5.95}\\
UCAM                &20.52$\pm$4.00&53.14$\pm$2.37          &41.02$\pm$4.96          &35.72$\pm$4.07          &44.67$\pm$5.47          &29.32$\pm$6.52          &27.38$\pm$6.42\\
online-Y            &20.46$\pm$3.42&\oosmark{53.72$\pm$1.79}&\oosmark{41.14$\pm$4.47}&\oosmark{37.22$\pm$4.83}&44.53$\pm$5.48          &\oosmark{29.65$\pm$6.33}&\oosmark{27.75$\pm$6.14}\\
dfki-nmt            &20.30$\pm$3.11&\oosmark{53.74$\pm$1.75}&40.96$\pm$4.18          &36.92$\pm$4.65          &43.67$\pm$4.81          &28.88$\pm$5.98          &26.97$\pm$5.78\\
MLLP-UPV            &20.30$\pm$3.47&53.45$\pm$2.00          &40.75$\pm$4.57          &36.75$\pm$4.49          &\oosmark{43.80$\pm$4.99}&28.81$\pm$6.12          &26.84$\pm$5.98\\
PROMT\_NMT          &20.16$\pm$2.88&53.27$\pm$1.26          &40.46$\pm$3.69          &36.41$\pm$4.84          &\oosmark{43.88$\pm$4.85}&28.76$\pm$5.44          &26.73$\pm$5.45\\
eTranslation        &20.12$\pm$3.47&\oosmark{53.45$\pm$2.00}&\oosmark{40.73$\pm$4.42}&36.22$\pm$4.26          &43.45$\pm$4.90          &28.17$\pm$5.89          &26.15$\pm$5.63\\
UdS-DFKI            &20.05$\pm$3.31&51.41$\pm$1.40          &39.39$\pm$3.89          &33.37$\pm$8.07          &\oosmark{45.36$\pm$4.99}&\oosmark{28.80$\pm$5.62}&\oosmark{26.97$\pm$5.59}\\
JHU                 &19.89$\pm$3.02&\oosmark{52.93$\pm$1.64}&\oosmark{40.53$\pm$4.23}&\oosmark{36.20$\pm$5.19}&44.09$\pm$4.83          &\oosmark{28.92$\pm$5.92}&26.95$\pm$5.81\\
TartuNLP-c          &19.67$\pm$3.33&52.72$\pm$1.31          &39.93$\pm$4.11          &36.15$\pm$5.01          &\oosmark{44.18$\pm$5.86}&28.56$\pm$6.11          &26.58$\pm$6.03\\
online-A            &19.36$\pm$3.71&52.47$\pm$2.15          &39.73$\pm$4.68          &34.63$\pm$3.52          &42.36$\pm$5.39          &27.17$\pm$6.66          &25.23$\pm$6.49\\
online-G            &18.80$\pm$3.41&52.26$\pm$1.35          &38.97$\pm$3.86          &\oosmark{34.89$\pm$4.93}&\oosmark{44.69$\pm$5.34}&\oosmark{28.53$\pm$5.84}&\oosmark{26.73$\pm$5.88}\\
online-X            &13.66$\pm$2.22&48.06$\pm$1.12          &33.85$\pm$3.67          &31.48$\pm$5.69          &30.69$\pm$5.80          &17.04$\pm$6.51          &15.42$\pm$6.24\\
en\_de\_task        &10.44$\pm$1.93&42.22$\pm$1.25          &28.15$\pm$2.92          &22.23$\pm$6.98          &\oosmark{34.90$\pm$5.05}&16.85$\pm$5.52          &15.15$\pm$5.50\\
Microsoft-sent\_doc &0.00$\pm$0.00 &0.12$\pm$0.02           &0.00$\pm$0.00           &-3408.43$\pm$471.13     &0.00$\pm$0.00           &0.00$\pm$0.00           &0.00$\pm$0.00\\

\end{tabular}
\end{center}
\caption{Automatic scores for English\to{}German. ``\oossymb'' marks scores out of sequence. }
\label{auto:en-de}
\end{table*}

\begin{table*}[t]
\begin{center}
\scriptsize
\begin{tabular}{lrrrrrrr}
              &BLEU          &chrF3                   &nCDER                   &nCharacTER              &nPER                    &nTER                    &nWER\\
\hline
Facebook\_FAIR&26.81$\pm$2.95&52.76$\pm$2.38          &46.17$\pm$3.07          &35.78$\pm$3.89          &57.82$\pm$2.70          &39.59$\pm$4.03          &36.73$\pm$4.04\\
RWTH\_Aachen  &26.02$\pm$3.01&51.74$\pm$2.52          &45.53$\pm$3.16          &35.61$\pm$3.66          &57.09$\pm$3.29          &39.16$\pm$4.29          &36.41$\pm$4.34\\
online-B      &25.62$\pm$3.06&51.30$\pm$2.57          &45.30$\pm$3.44          &33.97$\pm$4.02          &56.42$\pm$3.65          &\oosmark{39.70$\pm$4.30}&\oosmark{36.99$\pm$4.21}\\
NEU           &25.45$\pm$2.84&\oosmark{51.55$\pm$2.27}&45.19$\pm$2.85          &\oosmark{35.27$\pm$3.97}&\oosmark{57.04$\pm$3.10}&38.83$\pm$3.80          &36.09$\pm$3.81\\
online-Y      &25.27$\pm$3.26&51.30$\pm$2.40          &\oosmark{45.38$\pm$3.34}&35.01$\pm$3.58          &56.52$\pm$3.46          &\oosmark{39.77$\pm$4.20}&\oosmark{36.93$\pm$4.11}\\
dfki-nmt      &25.00$\pm$2.90&50.89$\pm$2.18          &44.64$\pm$2.89          &34.67$\pm$3.47          &56.21$\pm$3.18          &38.34$\pm$3.96          &35.65$\pm$3.94\\
UCAM          &24.95$\pm$3.37&50.44$\pm$2.60          &44.64$\pm$3.25          &33.83$\pm$4.38          &56.21$\pm$3.57          &38.30$\pm$4.03          &35.51$\pm$3.95\\
MSRA.MADL     &24.86$\pm$3.59&\oosmark{50.73$\pm$2.65}&44.38$\pm$3.32          &33.17$\pm$4.24          &55.73$\pm$4.53          &36.23$\pm$5.93          &33.37$\pm$5.83\\
JHU           &24.82$\pm$2.97&50.56$\pm$1.94          &44.38$\pm$2.84          &\oosmark{33.98$\pm$3.74}&\oosmark{55.92$\pm$2.83}&\oosmark{36.90$\pm$3.62}&\oosmark{34.14$\pm$3.61}\\
MLLP-UPV      &24.39$\pm$3.30&50.20$\pm$2.13          &44.20$\pm$3.07          &32.97$\pm$4.24          &55.91$\pm$3.18          &\oosmark{37.72$\pm$4.04}&\oosmark{34.89$\pm$3.98}\\
online-A      &24.13$\pm$3.41&50.03$\pm$2.57          &44.06$\pm$3.64          &32.95$\pm$3.73          &55.34$\pm$3.72          &\oosmark{37.87$\pm$4.62}&\oosmark{35.26$\pm$4.66}\\
online-G      &24.11$\pm$3.38&\oosmark{50.52$\pm$2.08}&43.80$\pm$3.07          &\oosmark{34.19$\pm$4.50}&\oosmark{55.55$\pm$3.10}&36.49$\pm$4.09          &33.75$\pm$4.10\\
TartuNLP-c    &23.82$\pm$2.80&50.46$\pm$2.27          &\oosmark{43.83$\pm$3.15}&33.30$\pm$3.31          &54.88$\pm$3.31          &\oosmark{38.45$\pm$3.50}&\oosmark{35.56$\pm$3.57}\\
PROMT\_NMT    &22.58$\pm$2.29&49.29$\pm$2.12          &42.48$\pm$2.46          &32.80$\pm$3.65          &53.98$\pm$2.95          &36.02$\pm$3.56          &33.31$\pm$3.31\\
uedin         &21.37$\pm$3.34&47.22$\pm$3.04          &41.30$\pm$3.68          &25.52$\pm$7.74          &50.68$\pm$4.07          &\oosmark{37.55$\pm$4.00}&\oosmark{35.16$\pm$3.87}\\
online-X      &17.95$\pm$2.09&44.93$\pm$2.26          &38.38$\pm$2.42          &\oosmark{26.69$\pm$4.34}&49.95$\pm$2.76          &32.69$\pm$3.06          &30.23$\pm$3.03\\

\end{tabular}
\end{center}
\caption{Automatic scores for German\to{}English. ``\oossymb'' marks scores out of sequence. }
\label{auto:de-en}
\end{table*}

\begin{table*}[t]
\begin{center}
\scriptsize
\begin{tabular}{lrrrrrrr}
             &BLEU          &chrF3                   &nCDER                   &nCharacTER              &nPER                    &nTER                    &nWER\\
\hline
online-B     &15.67$\pm$4.40&47.16$\pm$4.21          &33.60$\pm$5.83          &28.12$\pm$4.94          &42.24$\pm$5.92          &23.10$\pm$7.18          &21.22$\pm$6.96\\
online-Y     &15.55$\pm$4.20&\oosmark{47.71$\pm$3.97}&\oosmark{34.32$\pm$6.06}&\oosmark{31.75$\pm$5.20}&39.59$\pm$6.17          &21.96$\pm$7.47          &20.22$\pm$7.17\\
online-A     &13.15$\pm$3.38&45.45$\pm$3.65          &31.95$\pm$5.28          &27.51$\pm$4.77          &35.61$\pm$5.13          &18.19$\pm$6.39          &16.57$\pm$6.07\\
online-G     &12.69$\pm$3.25&45.36$\pm$3.34          &31.29$\pm$4.92          &\oosmark{28.96$\pm$4.62}&\oosmark{36.98$\pm$5.60}&\oosmark{18.80$\pm$6.67}&\oosmark{17.01$\pm$6.39}\\
\hline
NICT         &10.61$\pm$2.39&43.24$\pm$2.48          &29.49$\pm$4.24          &27.46$\pm$3.84          &27.13$\pm$4.75          &11.51$\pm$5.88          &10.04$\pm$5.58\\
NEU\_KingSoft&9.34$\pm$2.80 &40.09$\pm$2.04          &27.38$\pm$4.87          &22.78$\pm$3.86          &26.39$\pm$6.29          &10.11$\pm$7.21          &8.71$\pm$6.93\\
Nanjing      &6.85$\pm$2.15 &35.73$\pm$2.20          &24.02$\pm$4.17          &19.40$\pm$5.42          &23.37$\pm$5.10          &6.68$\pm$5.63           &5.41$\pm$5.24\\

\end{tabular}
\end{center}
\caption{Automatic scores for Czech\to{}German. ``\oossymb'' marks scores out of sequence. Note that online systems use parallel data
while the others use only monolingual data.}
\label{auto:cs-de}
\end{table*}

\begin{table*}[t]
\begin{center}
\scriptsize
\begin{tabular}{lrrrrrrr}
                          &BLEU          &chrF3                   &nCDER                   &nCharacTER              &nPER                    &nTER                    &nWER\\
\hline
online-B                  &14.86$\pm$4.01&40.69$\pm$2.96          &32.04$\pm$4.91          &22.32$\pm$5.07          &40.86$\pm$4.53          &26.12$\pm$7.74          &24.43$\pm$7.53\\
online-Y                  &14.69$\pm$3.82&40.68$\pm$2.92          &\oosmark{32.12$\pm$4.66}&\oosmark{24.69$\pm$4.88}&\oosmark{40.87$\pm$4.48}&26.02$\pm$7.20          &24.33$\pm$6.80\\
online-G                  &12.22$\pm$2.71&39.16$\pm$1.94          &29.59$\pm$3.44          &22.13$\pm$5.36          &38.75$\pm$3.86          &21.90$\pm$5.76          &20.32$\pm$5.48\\
online-A                  &11.80$\pm$2.92&38.09$\pm$2.52          &28.92$\pm$4.11          &21.11$\pm$5.35          &37.42$\pm$5.20          &\oosmark{22.17$\pm$7.38}&\oosmark{20.51$\pm$7.14}\\
\hline
NICT                      &10.49$\pm$2.95&35.99$\pm$3.00          &27.20$\pm$4.37          &20.08$\pm$5.78          &36.49$\pm$4.69          &19.63$\pm$6.55          &18.10$\pm$6.17\\
NEU\_KingSoft             &8.18$\pm$2.65 &32.89$\pm$2.86          &24.61$\pm$4.94          &16.94$\pm$5.84          &32.62$\pm$4.93          &19.61$\pm$7.08          &\oosmark{18.36$\pm$6.75}\\
lmu-unsup-nmt             &7.40$\pm$2.49 &31.69$\pm$2.46          &22.96$\pm$4.14          &13.86$\pm$4.88          &30.72$\pm$4.27          &18.00$\pm$6.07          &16.91$\pm$5.90\\
CUNI-Unsupervised-NER-post&7.03$\pm$2.26 &\oosmark{32.40$\pm$2.46}&22.76$\pm$4.18          &\oosmark{14.59$\pm$4.51}&\oosmark{31.46$\pm$4.42}&17.43$\pm$6.11          &16.13$\pm$5.77\\
Nanjing-6929              &6.26$\pm$2.11 &28.42$\pm$2.00          &21.11$\pm$3.89          &9.16$\pm$7.70           &28.55$\pm$4.02          &13.92$\pm$6.20          &13.00$\pm$6.10\\
Nanjing-6935              &6.26$\pm$2.11 &28.42$\pm$2.00          &21.11$\pm$3.89          &9.16$\pm$7.70           &28.55$\pm$4.02          &13.92$\pm$6.20          &13.00$\pm$6.10\\
CAiRE                     &5.85$\pm$2.05 &26.75$\pm$2.21          &20.13$\pm$3.41          &4.52$\pm$7.22           &\oosmark{29.14$\pm$4.50}&\oosmark{14.16$\pm$5.07}&\oosmark{13.03$\pm$4.74}\\

\end{tabular}
\end{center}
\caption{Automatic scores for German\to{}Czech. ``\oossymb'' marks scores out of sequence. Note that online systems use parallel data
while the others use only monolingual data.}
\label{auto:de-cs}
\end{table*}

The main observation across the tables is that all the scores heavily vary
across individual documents. The typical standard deviation is 3--5 for BLEU and
similarly for other metrics.

The metrics do not always agree on the overall ranking of the systems, as
indicated by ``\oossymb'' in the tables, but these differences are much smaller
that the variance due to the particular documents.

A big caveat should be taken when interpreting all automatic scores as an
estimate of real translation quality, because they are all based on the single
reference translation. See also the discussion in \cref{ref-problems} below.

\section{Manual Evaluation}
\label{sec:evaluation}

Due to the specific terminology in the documents and domain knowledge needed to
verify translation quality, we asked the SAO's employees serve as the annotators.\footnote{
Relying on purely linguistic expertise proved insufficient after a discussion with SAO employees. While for the best systems, we could hardly notice any errors, the knowledge experts discussed term choice and even among themselves, they were carefully considering the logical implications of the particular terms.}
All of them were native Czech speakers with a high level of English and/or German proficiency.

We also attempted to find native German auditors but we were not successful so far.
English\to{}German and German\to{}English translation was thus evaluated by a
single SAO employee, a native Czech speaker with a great command of both English
and German, including the specific auditing domain.

\subsection{Establishing Evaluation Criteria}
\label{sec:evaluation-criteria}

Our manual evaluation criteria are based on the criteria used for the scoring of
essays in the Czech GCSE counterpart (``maturita'') for the Czech language.



After a short test session with our prospective annotators, we realized how very
narrow this specific field is and we simplified the original set of 7 criteria
with 6 levels each to only 5 criteria and 4 levels each.
This simplification definitely saved some annotation time and we also believe that it increased the inter-annotator agreement, although we did not collect enough annotations to reliably measure it.
%

The final criteria to be used in the evaluation are as follows:

\paragraph{1) Language Resources -- Spelling and Morphology}
\begin{itemize}
    \item 0 points: 10 or more spelling or
	morphology errors.
    \item 1 point: 9-6 spelling or morphology errors.
    \item 2 points: 5-3 spelling or morphology errors.
    \item 3 points: 2-0 spelling or morphology errors.
\end{itemize}

\paragraph{2) Vocabulary -- Adequacy of Terms Used}
\begin{itemize}
    \item 0 points: Frequently, used terms are inappropriately chosen.
    \item 1 point:  Sometimes, used terms are inappropriately chosen.
    \item 2 points: Rarely, used terms are inappropriately chosen.
    \item 3 points: There are no terms, which would be inappropriately chosen.
\end{itemize}

\paragraph{3) Vocabulary -- Clarity of the Text in Terms of Used Words}
\begin{itemize}
    \item 0 points: The choice of words and phrases fundamentally impairs the understanding of the text.
    \item 1 point: The choice of words and phrases sometimes impairs the understanding of the text.
    \item 2 points: The choice of words and phrases rarely impairs the understanding of the text.
    \item 3 points: The choice of words and phrases does not impair the understanding of the text.
\end{itemize}

\paragraph{4) Syntax and Word Order}
\begin{itemize}
    \item 0 points: Syntactic shortcomings are high in the text.
    \item 1 point: Syntactic shortcomings occur in the text.
    \item 2 points: Syntactic shortcomings are rare in the text.
    \item 3 points: Syntactic shortcomings are almost absent from the text.
\end{itemize}

\paragraph{5) Coherence and Overall Understanding of the Text}
\begin{itemize}
    \item 0 points: The recipient is completely lost in the text. The text is incoherent and fails to fulfil its communication purpose (the addressee has completely misunderstood what the text expresses).
    \item 1 point: The orientation in the text is completely uncomfortable for the addressee, the text is at times incoherent and barely serves its communication purpose (but the addressee believes that he or she understands the main content of the text more or less).
    \item 2 points: The recipient navigates the text, though not entirely comfortably. The text is coherent and more or less fulfils its communication purpose (the addressee is sure he understands the text as a whole).
    \item 3 points: The recipient is fully oriented in the text. The text is completely coherent, it serves its communication purpose excellently (the addressee fully and without difficulty understands the text as a whole).
\end{itemize}

\subsection{Reference Effectively Useless}
\label{ref-problems}

One observation that emerged from our consultation with the experts in the auditing field
was that 
precise choice of terms is extremely important but that detailed knowledge of the
respective legislation and practice is necessary to evaluate the translations.
We, highly proficient speakers of English, but lacking any substantial
information on taxation and other topics discussed in the documents, often could
not see any lexical errors, because at the general level, the choice of words
seemed acceptable. The experts discussed at length the various factual
implications of using one of the near-synonyms over another.

Anecdotally,
\emph{voting} among our three consultants would not always work either. Without
a chance to discuss a particular term, two of the consultants would label the
choice of an MT system as wrong, but the third consultant, the most experienced
expert in the very field actually approved it.

The reference translations proved effectively useless for these fine
distinctions,
because the particular term used in the single reference was often not the only
possible one. As already mentioned, the careful revision applied to the
reference translations has sometimes slightly shifted the meaning, preferring a
better match with the factual knowledge over the literality of the translation.


\subsection{Execution of Evaluation}
\label{sec:execution}

As was mentioned above, the annotators were the employees of the SAO.

We decided to score not the complete documents but rather selected segments of about 15 consecutive sentences.
Each such segment takes something between a half and a full A4 page when printed.

For each evaluated page, the annotators were provided with another such page---the corresponding 15
sentences in the source language. We deliberately avoided providing reference
translations for two reasons: (1) we included the reference as if it was
one of the competing MT systems, (2) we know that the source and the
reference occasionally departed from each other; judging MT systems based on the
references would thus not be a fair comparison even if carried out by humans and
not an automatic metric.

In a small probe, we estimated that the annotation of one such segment will take
about 15 minutes.

\begin{table}[t]
\centering
\begin{tabular}{lrrr}
\textbf{Langs.} & \textbf{\# Doc Segments} & \textbf{\# Annotators}\\ 
\hline
en-cs & 48 & 5 \\
en-de & 16 & \multirow{2}{*}{1}   \\
de-en & 16 &    \\
\end{tabular}
\caption{Summary of manual annotations.}
\label{tab:anot:stats}
\end{table}

\cref{tab:anot:stats} summarizes the number of annotated document segments and
annotators providing the scores.


The actual evaluation of each segment was submitted by the annotators
through a
simple web interface, which recorded:
\begin{itemize}
    \item the segment ID;
    \item points assigned to the evaluated categories;
	\item a free-form description of the most serious error(s);
	\item a free-form field for further comments;
	\item a check-box indicating whether the annotator is an
    expert in the given field of the segment (e.g. in the field of value-added tax, VAT).
\end{itemize}

\subsection{Results of Manual Evaluation}
\label{results}

\def\manevaltab#1#2{
\begin{table*}[t]
\begin{center}
\scriptsize
\begin{tabular}{l rrrrr}
\input{manual-eval.#1.tex}
\end{tabular}
\end{center}
\caption{#2}
\label{tab:man-eval-#1}
\end{table*}
}

\begin{table*}[t]
\begin{center}
\scriptsize
\begin{tabular}{l rrrrr}
 & Spell. \& morpho. & Vocab. -- adequacy & Vocab. -- clarity & Syntax \& word order & Coher. \& overall underst. \\
\hline
Reference                 & 2.38$\pm$0.70           & 2.44$\pm$0.46           & 2.44$\pm$0.46           & 2.50$\pm$0.71           & 2.50$\pm$0.50 \\
online-B                  & \oosmark{2.50$\pm$0.67} & 2.40$\pm$0.49           & 2.20$\pm$0.75           & \oosmark{2.60$\pm$0.66} & 2.40$\pm$0.66 \\
CUNI-DocTransformer-T2T   & \oosmark{2.75$\pm$0.43} & 2.25$\pm$0.83           & \oosmark{2.33$\pm$0.75} & 2.58$\pm$0.49           & 2.33$\pm$0.85 \\
CUNI-Transformer-T2T-2019 & 2.60$\pm$0.49           & \oosmark{2.50$\pm$0.67} & 2.30$\pm$0.78           & 2.40$\pm$0.49           & 2.30$\pm$0.78 \\
TartuNLP-c                & 1.88$\pm$0.78           & 1.62$\pm$0.86           & 1.75$\pm$0.83           & 1.88$\pm$0.93           & 1.75$\pm$0.97 \\

\end{tabular}
\end{center}
\caption{Mean scores of English\to{}Czech translation obtained in manual evaluation. The systems are sorted by the ``coherence and overall understanding'' criterion.
Higher scores are better.
``\oossymb'' marks scores out of sequence.}
\label{tab:man-eval-en-cs-abs}
\end{table*}

\begin{table*}[t]
\begin{center}
\scriptsize
\begin{tabular}{l rrrrr}
 & Spell. \& morpho. & Vocab. -- adequacy & Vocab. -- clarity & Syntax \& word order & Coher. \& overall underst. \\
\hline
online-B                  & 1.80$\pm$0.98           & 1.60$\pm$0.80           & 2.00$\pm$1.10           & 1.40$\pm$0.80 & 1.80$\pm$0.75 \\
Reference                 & 2.75$\pm$1.09           & 1.75$\pm$0.83           & \oosmark{1.75$\pm$0.83} & 2.00$\pm$1.22 & 2.00$\pm$0.71 \\
CUNI-DocTransformer-T2T   & \oosmark{1.40$\pm$0.80} & 2.60$\pm$1.62           & 2.00$\pm$1.55           & 2.20$\pm$0.75 & 2.20$\pm$1.47 \\
CUNI-Transformer-T2T-2019 & 1.75$\pm$0.83           & \oosmark{2.00$\pm$1.00} & 2.50$\pm$1.12           & 2.75$\pm$1.48 & 2.25$\pm$1.09 \\
TartuNLP-c                & 3.40$\pm$1.96           & 4.00$\pm$0.63           & 3.00$\pm$0.89           & 3.00$\pm$1.41 & 3.20$\pm$1.47 \\

\end{tabular}
\end{center}
\caption{Mean ordinal numbers of English\to{}Czech systems sorted by manual evaluation scores for each annotator.
Lower numbers are better.
}
\label{tab:man-eval-en-cs-rel}
\end{table*}


We did not have enough human capacity to calculate an full-fledged inter-annotator agreement.
To have at least some idea of how annotators agree, we let three of all segments be assessed by two different annotators.
Comparison of the scores reveals that annotators often differ in their assessment, even though the assigned points are almost always neighbouring.

Somewhat surprisingly, except for a single segment, the annotators did not consider
themselves experts in the field of the documents presented to them, even though
they all should be professionals in the auditing field.

\subsubsection{English-to-Czech Translation}

Altogether, the English\to{}Czech translations were evaluated by 5 annotators.
They evaluated 48 segments randomly chosen from documents translated by 4 selected systems and the reference translation.
The translation systems were selected based on their automatic scores in WMT19 and their results in the past years.
TartuNLP-c  was added as a representative of a system with an overall lower
output quality, although it seemed to perform well in some of the observed phenomena.

\cref{tab:man-eval-en-cs-abs} shows the mean scores and standard deviations
collected on the translations according to the five criteria specified in
\cref{sec:evaluation-criteria}.

As our mini-comparison of annotator agreement suggests mismatches in score
assignments, we provide also a statistic that abstracts from the absolute values
of assigned scores. Because the assigned scores are associated with a particular
categorical description, we avoid the standard normalization of mean and
variance. Instead,
we take all the assessments produced by a single annotator and sort the systems
by the average of scores assigned by him or her in a given criterion.
\cref{tab:man-eval-en-cs-rel} then shows the mean ordinal number of each of the systems across all the annotators.
Unlike the scores in \cref{tab:man-eval-en-cs-abs}, the best ordinal number is 1 and it  gets worse as it increases.

\begin{table*}[t]
\begin{center}
\scriptsize
\begin{tabular}{l rrrrr}
 & Spell. \& morpho. & Vocab. -- adequacy & Vocab. -- clarity & Syntax \& word order & Coher. \& overall underst. \\
\hline
Reference    & 3.00$\pm$0.00           & 3.00$\pm$0.00 & 3.00$\pm$0.00 & 3.00$\pm$0.00 & 3.00$\pm$0.00 \\
MSRA-MADL    & 2.75$\pm$0.43           & 2.25$\pm$0.83 & 2.25$\pm$0.83 & 2.25$\pm$0.83 & 2.25$\pm$0.83 \\
eTranslation & 2.50$\pm$0.50           & 2.25$\pm$0.83 & 2.25$\pm$0.83 & 2.25$\pm$0.83 & 2.00$\pm$1.00 \\
online-B     & \oosmark{2.75$\pm$0.43} & 1.75$\pm$1.30 & 1.75$\pm$1.30 & 2.00$\pm$0.71 & 1.50$\pm$1.12 \\

\end{tabular}
\end{center}
\caption{Mean scores of English\to{}German translation obtained in
manual evaluation. The systems are sorted by the ``coherence and overall
understanding'' criterion. Higher scores are better.}
\label{tab:man-eval-en-de-abs}
\end{table*}

\begin{table*}[t]
\begin{center}
\scriptsize
\begin{tabular}{l rrrrr}
 & Spell. \& morpho. & Vocab. -- adequacy & Vocab. -- clarity & Syntax \& word order & Coher. \& overall underst. \\
\hline
Reference & 2.60$\pm$0.49           & 2.60$\pm$0.49 & 2.60$\pm$0.49 & 2.60$\pm$0.49 & 2.60$\pm$0.49 \\
online-B  & 2.33$\pm$0.47           & 2.17$\pm$0.69 & 1.83$\pm$0.69 & 2.00$\pm$0.58 & 1.83$\pm$0.69 \\
MSRA-MADL & \oosmark{2.40$\pm$0.49} & 1.60$\pm$0.80 & 1.60$\pm$0.80 & 1.80$\pm$0.75 & 1.60$\pm$0.80 \\

\end{tabular}
\end{center}
\caption{Mean scores of German\to{}English translation obtained in
manual evaluation. The systems are sorted by the ``coherence and overall
understanding'' criterion. Higher scores are better.}
\label{tab:man-eval-de-en-abs}
\end{table*}

Even though some subtle differences occur in ordering of the systems in Tables~\ref{tab:man-eval-en-cs-abs} and \ref{tab:man-eval-en-cs-rel}, the main observations remain the same.
Manual evaluation confirms the lower quality of TartuNLP-c measured by automatic metrics.
On the other hand, online-B scored best and it appears on par with the human
translation, whereas it was surpassed by CUNI systems in terms of the automatic
metrics as well as in news translation (see the main Findings of WMT19 paper).
Interestingly, apart from TartuNLP-c all the other MT systems seem to yield
fewer spelling and morphology errors than the human translators, although the
differences are within the standard deviation bounds. CUNI-DocTransformer-T2T
stands out by being better even beyond the reported standard deviation of the
ordinal interpretation (see 1.40$\pm$0.80 in ``Spell. \& morpho.'' in
\cref{tab:man-eval-en-cs-rel}).

Due to large values of standard deviations, the small sample size and the fact
that the underlying set of evaluated document segments varied across the systems,
it is difficult to draw reliable conclusions from these observations. Some
counter-intuitive results can be thus attributed to pure randomness. For
example, CUNI-Transformer-T2T-2019 differs from CUNI-DocTransformer-T2T only in
the fact that it operates on triples of consecutive sentences. This should
increase the adequacy of vocabulary chosen and should have no effect on spelling
and morphology but we have seen the opposite.

The overall statement we \emph{can} make is that for English-to-Czech, the specific domain of audit
reports does not differ much from the general observations made in the main News
Translation Task: the order of the systems generally matches and the better
systems are very close to the human performance.

\subsubsection{English$\leftrightarrow$German Translation}

Manual evaluation of English\to{}German translations was provided by a single annotator on 16 randomly selected segments, covering 3 systems and the human translation.
In the opposite translation direction, also 16 segments were evaluated by the same annotator, this time covering 2 systems and the human translation.
We chose the systems which are popular (online-B), expected to score among the best based
on their (automatically assessed) performance on the News Translation Task
(MSRA-MADL) or are provided by the European Commission as a service for EU
institutions (eTranslation).

The mean scores in Tables~\ref{tab:man-eval-en-de-abs} and
\ref{tab:man-eval-de-en-abs} show that none of the systems outperforms human
translation.
The ordering of the systems remains the same across most of the evaluation criteria.
Unlike in automatic evaluation, the human annotator considers the output of
online-B in English\to{}German translation of lower quality (except spelling and
morphology) than the outputs of
its competitors.
In German\to{}English translation, the ordering of the systems according to the manual evaluation agrees with the automatic one.

All in all, comparison of manual and automatic evaluation suggests  that the systems achieving high automatic scores may be judged differently by human annotators.
As the quality of translation decreases, it is sufficient to evaluate it automatically.

\subsubsection{Most Common Mistakes}

\begin{table}[t]
\centering
\begin{tabular}{lrrr}
\textbf{Errors in} & \textbf{en-cs} & \textbf{en-de}  & \textbf{de-en} \\ 
\hline
Wrong translation   &  20 &  14 & 28  \\
Fluency             &  25 &   1 & 0 \\
Untranslated        &   5 &   3 & 7  \\
Abbreviations       &   6 &   4 & 4  \\
Grammar             &   8 &   2 & 2  \\
Missing words       &   4 &   0 & 2  \\
Coherence           &   4 &   1 & 0  \\
Added words         &   4 &   0 & 0  \\
Word repetition     &   2 &   0 & 2  \\
Spasm               &   1 &   0 & 0  \\
\hline
Total               &  79 &  25 & 45 \\
Avg. per Doc. Segm. & 1.6 & 1.5 &    2.8
\end{tabular}
\caption{Summary of errors found by SAO annotators.}
\label{tab:errors:summary}
\end{table}

A part of the evaluation web interface was a free-form field for the description of
the most serious error(s) encountered. We collected these comments and manually
organized them into several categories.
We found out that the most common mistakes were:

\begin{itemize}
    \item fluency;
    \item wrong translation of terms;
    \item grammatical correctness (such as a wrong gender chosen for pronouns);
    \item non-translated abbreviations, or abbreviations which do not make sense in the Czech translation;
    \item outputs completely missing a half of the sentence. This was particularly likely after a punctuations such as the closing bracket in the middle of the sentence.
\end{itemize}


\cref{tab:errors:summary} summarizes the overall error counts by category.
(The reference is included in these counts.)
As mentioned above, we did not find any native German auditor who could annotate our SAO Test Suite, so the annotation was done by a single Czech auditor. This could explain the relatively big differences between language pairs: with a single annotation, the annotator disagreements are not averaged out.
For instance, it is possible that this marked some of the errors as wrong grammatical constructions while en\to{}cs annotators could score it in fluency criterion.

We also have to take into account the absolute number of annotated document
segments (48 for Czech, 16 for English$\leftrightarrow$German). Considering the
average number of errors per one annotated document segment, German\to{}English
translation seems the worst, see the last line of \cref{tab:errors:summary}.


\section{Translation of Agreements}
\label{sec:sml}

Aside from the SAO audit documents, we added one moderately long document from a very specific domain related to auditing: agreements.

As the source document, we used the English version of a sublease agreement, which was in fact a (non-professional) translation from Czech.
The original Czech text was evaluated with all other WMT19 systems as if it was one of the systems.

Due to the different nature of the text, we decided to evaluate the translation
of the sublease agreement differently from the evaluation of the main part of
SAO Test Suite.

\subsection{Manual Evaluation}
\label{sml:manual}
The evaluation of this small set containing one source document, one human
translation and 11 machine translated documents was done manually.
The evaluation was partially blind.
Technically, the candidate translations
were not labelled with the system name, but the main annotator could guess some of
the systems. Only the systems online-X, Y and G are truly blind, we do not know
their identity even from past evaluations.

We are confident that even the knowledge of the MT system did not affect our
evaluation because we fully focused on the hard criteria such as named entity preservation or term consistence throughout the document.
The only soft criterion included was the ``fluency'' one.
We have also included the reference document in the evaluation. 

\subsection{Establishing Evaluation Criteria}

By inspecting several of the MT outputs, we first defined the assessment criteria.
They generally fall into two categories: (1) target-only, and (2) source-based.
Whereas in the former category, we consider only quality of the target texts on
their own, regardless the source, in the latter we validate if the selected bits
of information were preserved or corrupted during the translation process.

In the target-only category, we focused on the following:
 \begin{itemize}
     \item fluency;
     \item grammatical correctness (this is very strict and well defined in
	 Czech; most errors were in morphological agreement and sometimes
	 verb tense);
     \item casing errors (esp. in named entities);
     \item incomprehensibility of the segment;
     \item ``spasm'', i.e. the situation when the MT system gets stuck in repeating some tokens;
     \item superfluous words;
     \item missing words or a whole sentence.
 \end{itemize}

\begin{figure*}[t]
\begin{center}
\includegraphics[width=.45\textwidth]{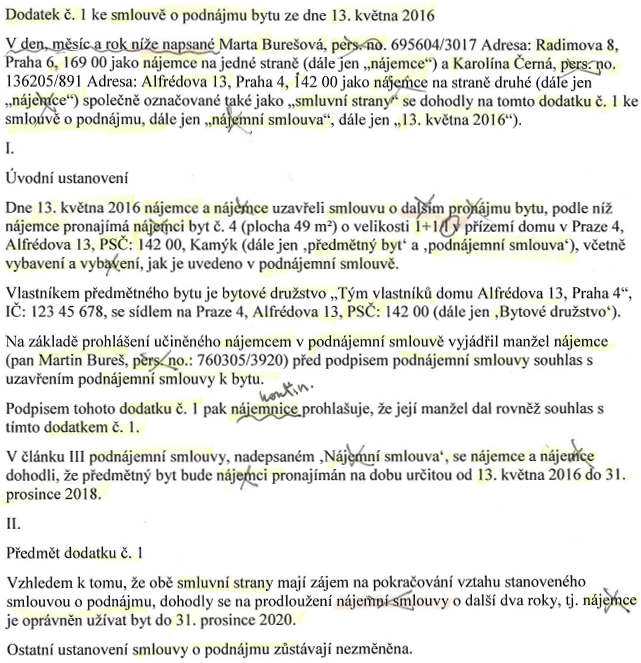}
\includegraphics[width=.45\textwidth]{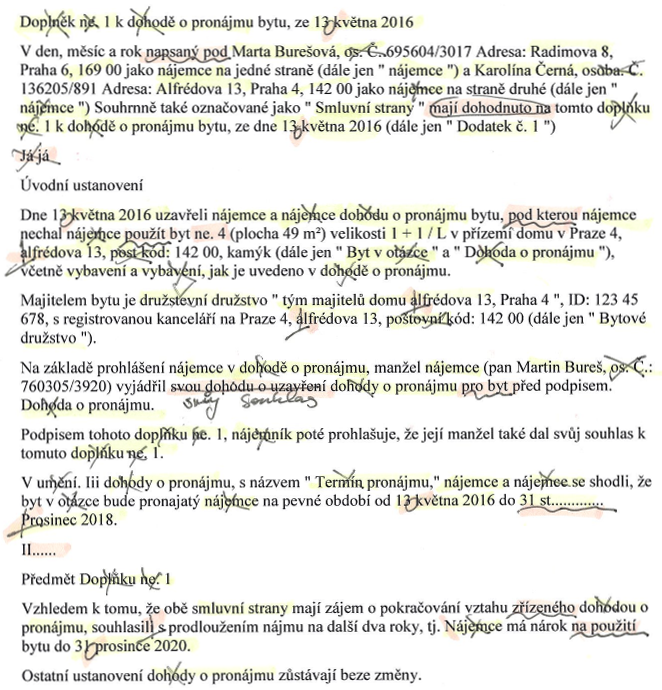}
\end{center}
\caption{Samples from our annotation with one of the best scoring systems
(CUNI-Transformer-T2T-2018) on the left and one of the worst ones (online-X) on
the right. Crosses indicate errors in term translation, strange wordings are
underlined, casing errors and other errors have their simple marks.}
\label{skeny}
\end{figure*}

As for the source-based category, we have focused on the errors, which were
formed either by wrong translation of a very domain-specific term or an
inconsistence of used terms throughout the whole document.
 \begin{itemize}
     \item Named Entities---here we checked mainly the preservation of the information:
     \begin{itemize}
         \item Person (e.g. name and surname);
         \item Address (e.g. street name and number);
         \item Date (esp. whether the format has been kept consistent);
         \item Numbers (if the transcription of numerals was correct);
         \item Flat composition (the Czech-specific way is to count rooms and
		 kitchens/kitchinette and indicate it as a compact string, here ``1+1'');
         \item Wrong abbreviation;
         \item Expanded abbreviation (e.g. in Czech, the ``ZIP CODE'' should be
		 translated as ``PS\v{C}'', which stands for ``po\v{s}tovn\'{i}
		 sm\v{e}rovac\'{i} \v{c}\'{i}slo'', but this abbreviation is never
		 spelled out in written text).
     \end{itemize}
     \item Document-specific terms:
     \begin{itemize}
         \item Tenant;
         \item Lessee;
         \item Supplement (of the agreement);
         \item Sublease agreement;
         \item Contracting parties;
         \item Apartment in question;
         \item Equipment (e.g. the kitchen);
         \item Amenities (e.g. a cellar or a segment of the garden);
         \item Housing cooperative;
         \item Team of owners;
         \item Term of the lease;
         \item The specification of the supplement (``no.~1'').
     \end{itemize}
 \end{itemize}

In the category of ``Document-specific terms'', we focused on evaluation
whether:
\begin{itemize}
    \item the term is translated correctly, incorrectly (incl. not translated
	at all), or missing altogether;
    \item the target term is preserved in the document.
\end{itemize}

It should be noted that the MT system was often free to choose from
several translation options of a term. 
At the same time, a very important criterion was whether the translation of each
of the terms was consistent throughout the document and also whether it did not
clash with other choices. For example, each of the terms ``tenant'' and ``lessee''
could be---depending on the particular situation---correctly translated as
``pronajímatelka'', ''nájemkyně'' or ``podnájemkyně''
(all are feminine variants
of the words, because incidentally, it was women who were entering this sample
agreement). If the two different parties however happened to have been referred
to in
any way that could lead to confusion, we marked this as a (serious) error.

In some cases, we had a strict expectation. For instance the term ``sublease''
could be translated into Czech in principle either as
``pronájem'' (which corresponds to the relationship between a landlord and a tenant)
or as ``podnájem'' (which corresponds to the relationship between a tenant and
a lessee). Based on the text of the agreement, it was however clear that the
correct term is ``podnájem'' (the tenant is not the actual owner of the
property), so we demanded the this particular choice.

\subsection{Execution of Evaluation}

Because of the relatively  small amount of data, the evaluation was done on
paper, see \cref{skeny}.

The annotations of ``source-based'' error types were done with respect to the
source text using a fixed set of ``markables'', i.e. the set of occurrences of words and expressions to
annotate for correctness. The set of markables was identical for all the candidate translations.
Each markable in each translation candidate received a label indicating if it was translated correctly,
with an error, or if was fully missing.

The ``target-only'' error types were marked independently for each system, with no number of markable positions given apriori.

\begin{table*}[t]
\centering
\begin{tabular}{lrrrrrrr}
 & \multicolumn{2}{c}{\bf Target-Only} & \multicolumn{2}{c}{\bf Source-Based} & \multicolumn{2}{c}{\bf Total} \\
\textbf{System} & \textbf{Errs} & \textbf{(Miss)} & \textbf{Errs} & \textbf{(Miss)} & \textbf{Errs} & \textbf{(Miss)}\\ 
\hline
Reference & 3 & 1 & 6 & 2 & 9 & 3\\
\TrEighteen{} & 6 & 0 & 15 & 0 & 21 & 0\\
\docTr{} & 9 & 0 & 21 & 0 & 30 & 0\\
\onlY{}& 10 & 0 & 20 & 0 & 30 & 0\\
\TrNineteen{} & 5 & 2 & 26 & 2 & 31 & 4\\
\onlB{} & 15 & 1 & 27 & 0 & 42 & 1\\
\uedin{} & 9 & 2 & 34 & 12 & 43 & 14\\
\onlA{} & 19 & 0 & 30 & 0 & 49 & 0 \\
\TrMarian{} & 13 & 2 & 38 & 0 & 51 & 2\\
\tartu{} & 14 & 1 & 37 & 1 & 51 & 2\\
\onlG{} & 34 & 0 & 28 & 0 & 62 & 0 \\
\onlX{} & 48 & 7 & 77 & 0 & 125 & 7\\
\end{tabular}
\caption{Total number of errors ``Errs'', and of those the cases when the output was completely missing ``(Miss)'', by English-Czech WMT19 news
translation systems applied to the sublease agreement.}
\label{tab:results}
\end{table*}

The question was how to deal with inconsistency in used terms. At the beginning it was not clear whether we should assume that the first occurrence of term ``defines'' it for the rest of the document or whether we should take the most frequent one as the ``intended one'' by the MT system and treat other translations as errors. After the first round of corrections, we chose the first option. Some terms, e.g. ``tenant'', ``lessee'' or ``agreement'' had always only one correct translation, but some, e.g. ``sublease'' could have had multiple possible translations. In these latter cases, we always marked the first occurrence as correct.

\subsection{Results of Manual Evaluation}

The summary of manual evaluation is presented in \cref{tab:results}. Errors in the source-based categories are more frequent than in target-only. 
This is mainly due to the incorrect translation of the term ``lessee'' (see \cref{lessee} below).

One thing worth mentioning is the 9 errors and 3 omissions in the reference
translation. 
This can be partly attributed to Czech being in fact the original and English (i.e. the source for MT systems) its translation. What is a good Czech\to{}English manual translation is not always literal enough when observed from the English side. Three errors were for instance incurred from one single case where the Czech text referred to the agreement itself one time less than the English text, but this ``missing reference'' (fully acceptable in the Czech\to{}English direction) counted as several missing expressions.
%
As for the true errors, there was one incorrect translation of term ``lessee'' and one mistake in the number of the Supplement.

  

The number of errors considerably varies across the systems. The best system
(CUNI-Transformer-T2T-2018) in our evaluation is also the winner on news in the
evaluation last year. As \perscite{findings:2018} report, this system
significantly outperformed humans \emph{at the level of individual sentences} in
that evaluation. In our setting, the number of errors by
CUNI-Transformer-T2T-2018 is twice the number of errors in the reference, but aside from term choice discussed in \cref{lessee}, one could say that the translation is very good.

In the target-only category, we did not have any pre-defined items that could be correct or incorrect. Therefore the number of errors varies greatly across the systems. From the lowest number of errors in the CUNI-Transformer-T2T-2019 (5 errors) and in CUNI-Transformer-T2T-2018 (6 errors) to the very high numbers in online-X and online-G (48 and 34 errors, respectively).

\def\blgrwh#1#2{
\begin{tikzpicture}[x=1em, y=1em]
\draw [use as bounding box] (0,0) rectangle (1,1);
\fill [fill=black] (0,0) rectangle (1, #1);
\fill [fill=gray] (0,#1) rectangle (1, #1+#2);
\end{tikzpicture}
}
\def\rot#1{
\begin{sideways}
#1
\end{sideways}
}

\begin{table*}
\begin{center}
\begin{tabular}{lr@{~}r@{~}r@{~}r@{~}r@{~}r@{~}r@{~}r@{~}r@{~}r@{~}r@{~}r@{~}r@{~}r@{~}r@{~}r@{~}r@{~}r@{~}r@{~}r@{~}r@{~}r@{~}r@{~}r@{~}r@{~}r@{~}r@{~}r@{~}r@{~}r@{~}r@{~}r}
              & \rot{Wrong Abbrev.} & \rot{Expanded Abbrev.} & \rot{Name}      & \rot{Surname}   & \rot{Street Name} & \rot{Number}    & \rot{Date}               & \rot{Apartment Specs} & \rot{Tenant}                        & \rot{Lessee}          & \rot{Supplement}       & \rot{Supplement Specs.} & \rot{Sublease Agreement}    & \rot{Contracting Parties}                   & \rot{Apartement in Question} & \rot{Equipment} & \rot{Amenities} & \rot{Housing Cooperative} & \rot{Team of Owners} & \rot{Term of the Lease} \\
Reference     & \blgrwh{0}{0}       & \blgrwh{0}{0}          & \blgrwh{0}{0}   & \blgrwh{0}{0}   & \blgrwh{0}{0}     & \blgrwh{0}{0}   & \blgrwh{0}{0}            & \blgrwh{0}{0}         & \blgrwh{0}{0}                       & \blgrwh{0.125}{0}     & \blgrwh{0}{0.0625}     & \blgrwh{0.125}{0.0625}  & \blgrwh{0}{0}               & \blgrwh{0}{0}                               & \blgrwh{0}{0}                & \blgrwh{0}{0}   & \blgrwh{0}{0}   & \blgrwh{0}{0}             & \blgrwh{0}{0}        & \blgrwh{0}{0} \\
\TrEighteen{} & \blgrwh{0}{0}       & \blgrwh{0}{0}          & \blgrwh{0}{0}   & \blgrwh{0}{0}   & \blgrwh{0}{0}     & \blgrwh{0}{0}   & \blgrwh{0}{0}            & \blgrwh{0}{0}         & \blgrwh{0.1111111111}{0}            & \blgrwh{1}{0}         & \blgrwh{0.0625}{0}     & \blgrwh{0}{0}           & \blgrwh{0.1666666667}{0}    & \blgrwh{0}{0}                               & \blgrwh{0}{0}                & \blgrwh{0}{0}   & \blgrwh{1}{0}   & \blgrwh{0}{0}             & \blgrwh{0}{0}        & \blgrwh{1}{0} \\
\docTr{}      & \blgrwh{1}{0}       & \blgrwh{0}{0}          & \blgrwh{0}{0}   & \blgrwh{0}{0}   & \blgrwh{0}{0}     & \blgrwh{0}{0}   & \blgrwh{0}{0}            & \blgrwh{0}{0}         & \blgrwh{0.1111111111}{0}            & \blgrwh{1}{0}         & \blgrwh{0}{0}          & \blgrwh{0}{0}           & \blgrwh{0.4166666667}{0}    & \blgrwh{0}{0}                               & \blgrwh{0}{0}                & \blgrwh{0}{0}   & \blgrwh{1}{0}   & \blgrwh{0}{0}             & \blgrwh{0}{0}        & \blgrwh{1}{0} \\
\onlY{}       & \blgrwh{1}{0}       & \blgrwh{0}{0}          & \blgrwh{0}{0}   & \blgrwh{0}{0}   & \blgrwh{0}{0}     & \blgrwh{0}{0}   & \blgrwh{0}{0}            & \blgrwh{0}{0}         & \blgrwh{0.1111111111}{0}            & \blgrwh{1}{0}         & \blgrwh{0.0625}{0}     & \blgrwh{0}{0}           & \blgrwh{0.3333333333}{0}    & \blgrwh{0}{0}                               & \blgrwh{1}{0}                & \blgrwh{0}{0}   & \blgrwh{1}{0}   & \blgrwh{0}{0}             & \blgrwh{0}{0}        & \blgrwh{1}{0} \\
\TrNineteen{} & \blgrwh{1}{0}       & \blgrwh{0}{0}          & \blgrwh{0}{0}   & \blgrwh{0}{0}   & \blgrwh{0}{0}     & \blgrwh{0}{0}   & \blgrwh{0}{0}            & \blgrwh{1}{0}         & \blgrwh{0.1111111111}{0.1111111111} & \blgrwh{0.875}{0.125} & \blgrwh{0.0625}{0}     & \blgrwh{0}{0}           & \blgrwh{0.6666666667}{0}    & \blgrwh{0}{0}                               & \blgrwh{1}{0}                & \blgrwh{0}{0}   & \blgrwh{1}{0}   & \blgrwh{0}{0}             & \blgrwh{0}{0}        & \blgrwh{1}{0} \\
\onlB{}       & \blgrwh{1}{0}       & \blgrwh{0.25}{0}       & \blgrwh{0}{0}   & \blgrwh{0}{0}   & \blgrwh{0}{0}     & \blgrwh{0}{0}   & \blgrwh{0}{0}            & \blgrwh{0}{0}         & \blgrwh{0}{0}                       & \blgrwh{1}{0}         & \blgrwh{0.3125}{0}     & \blgrwh{0}{0}           & \blgrwh{0.6666666667}{0}    & \blgrwh{0}{0}                               & \blgrwh{1}{0}                & \blgrwh{0}{0}   & \blgrwh{1}{0}   & \blgrwh{0}{0}             & \blgrwh{0}{0}        & \blgrwh{0}{0} \\
\uedin{}      & \blgrwh{1}{0}       & \blgrwh{0}{0}          & \blgrwh{0}{0}   & \blgrwh{0}{0}   & \blgrwh{0}{0}     & \blgrwh{0}{0}   & \blgrwh{0.1428571429}{0} & \blgrwh{1}{0}         & \blgrwh{0.2222222222}{0}            & \blgrwh{0.75}{0.25}   & \blgrwh{0.3125}{0.125} & \blgrwh{0}{0.125}       & \blgrwh{0.25}{0.1666666667} & \blgrwh{-3.5527136788005e-17}{0.1666666667} & \blgrwh{1}{0}                & \blgrwh{0}{1}   & \blgrwh{0}{1}   & \blgrwh{0}{0}             & \blgrwh{0}{0}        & \blgrwh{1}{0} \\
\onlA{}       & \blgrwh{1}{0}       & \blgrwh{0}{0}          & \blgrwh{0}{0}   & \blgrwh{0}{0}   & \blgrwh{0}{0}     & \blgrwh{0}{0}   & \blgrwh{0.1428571429}{0} & \blgrwh{0}{0}         & \blgrwh{0.2222222222}{0}            & \blgrwh{1}{0}         & \blgrwh{0.0625}{0}     & \blgrwh{0}{0}           & \blgrwh{1}{0}               & \blgrwh{0}{0}                               & \blgrwh{1}{0}                & \blgrwh{0}{0}   & \blgrwh{1}{0}   & \blgrwh{0.5}{0}           & \blgrwh{0}{0}        & \blgrwh{0}{0} \\
\TrMarian{}   & \blgrwh{1}{0}       & \blgrwh{0}{0}          & \blgrwh{0.4}{0} & \blgrwh{0.4}{0} & \blgrwh{0.4}{0}   & \blgrwh{0.2}{0} & \blgrwh{0.1428571429}{0} & \blgrwh{1}{0}         & \blgrwh{0.2222222222}{0}            & \blgrwh{1}{0}         & \blgrwh{0.0625}{0}     & \blgrwh{0}{0}           & \blgrwh{0.6666666667}{0}    & \blgrwh{0.1666666667}{0}                    & \blgrwh{0}{0}                & \blgrwh{0}{0}   & \blgrwh{1}{0}   & \blgrwh{0.5}{0}           & \blgrwh{0}{0}        & \blgrwh{1}{0} \\
\tartu{}      & \blgrwh{1}{0}       & \blgrwh{0.5}{0}        & \blgrwh{0}{0}   & \blgrwh{0}{0}   & \blgrwh{0}{0}     & \blgrwh{0}{0}   & \blgrwh{0}{0}            & \blgrwh{0}{0}         & \blgrwh{0.1111111111}{0}            & \blgrwh{0.875}{0.125} & \blgrwh{0.375}{0}      & \blgrwh{0}{0}           & \blgrwh{1}{0}               & \blgrwh{0}{0}                               & \blgrwh{1}{0}                & \blgrwh{0}{0}   & \blgrwh{1}{0}   & \blgrwh{0}{0}             & \blgrwh{0}{0}        & \blgrwh{1}{0} \\
\onlG{}       & \blgrwh{1}{0}       & \blgrwh{0}{0}          & \blgrwh{0}{0}   & \blgrwh{0}{0}   & \blgrwh{0}{0}     & \blgrwh{0}{0}   & \blgrwh{0}{0}            & \blgrwh{1}{0}         & \blgrwh{0.3333333333}{0}            & \blgrwh{1}{0}         & \blgrwh{0.0625}{0}     & \blgrwh{0}{0}           & \blgrwh{0.5}{0}             & \blgrwh{0}{0}                               & \blgrwh{1}{0}                & \blgrwh{1}{0}   & \blgrwh{1}{0}   & \blgrwh{0}{0}             & \blgrwh{1}{0}        & \blgrwh{1}{0} \\
\onlX{}       & \blgrwh{1}{0}       & \blgrwh{0.5}{0}        & \blgrwh{0}{0}   & \blgrwh{0}{0}   & \blgrwh{0}{0}     & \blgrwh{0}{0}   & \blgrwh{0.1428571429}{0} & \blgrwh{1}{0}         & \blgrwh{0.2222222222}{0}            & \blgrwh{1}{0}         & \blgrwh{0.9375}{0}     & \blgrwh{0.9375}{0}      & \blgrwh{1}{0}               & \blgrwh{0}{0}                               & \blgrwh{1}{0}                & \blgrwh{0}{0}   & \blgrwh{1}{0}   & \blgrwh{0.5}{0}           & \blgrwh{0}{0}        & \blgrwh{0}{0} \\

\end{tabular}
\end{center}
\caption{Composition of source-based errors of individual MT systems. An empty
box (\blgrwh{0}{0}) indicates no error. Black-filled portion corresponds to
erroneous output and gray-filled output corresponds to missing output.
}
\label{detailtable}
\end{table*}

As for the ``(Miss)'' counts, there were two types of situations: (1) only a single word was missing in the output and (2) the whole sentence or a half of a paragraph was not there. The second case often lead to a large increase in the ``(Miss)'' count because several markables from the source were supposed to appear in the lost part. The systems uedin and online-X were most affected by this.

Another interesting fact worth mentioning is that even though the system online-Y had a relatively low number of mistakes, those errors made the readability and the comprehensibility of the message substantially more difficult than e.g. the translation by online-B with a higher error count.

The point here is that the number of errors is important but their type can be critical, too.
We already mentioned the missing sentences or ``spasm'', which accounted for the 14 missing term translations in the output of uedin. Another interesting case is a ``misunderstanding'' of the MT system. For instance, {uedin} system misunderstood ``I.'' (the Roman numeral) for the pronoun ``I'' or mistranslated the ``ZIP CODE'' as ``občanka'' (personal ID card). It is exactly these types of errors, which are the most serious from the reader's point of view.


\subsubsection{Detailed Error Counts}

\cref{detailtable} provides further details on error types observed in the outputs of individual MT systems. The table is again sorted by the total number of errors as in \cref{tab:results}. We see that the best system (CUNI-Transformer-T2T-2018) fully failed in the translation of the terms ``lessee'', ``amenities'' and ``term of the lease''. This system was also the only one which dealt well with abbreviations.

In contrast to all other systems, CUNI-DocTransformer-Marian struggled to
translate several named entities correctly. This system used the same training
data as CUNI-Transformer-T2T-2019 and both of these systems translate several
consecutive sentences at once in order to improve cross-sentence consistency but
they somewhat differ in the details of the handling of multi-sentence input, and
they also differ in the underlying MT system: Tensor2Tensor vs. Marian, see
\perscite{cuni-news-wmt19} for more details. It is hard to explain why these
sentences could adversely affect named entities, so the authors of the system should carefully look at this issue.

\subsubsection{Referring to Contracting Parties}
\label{lessee}

\begin{table}[t]
    \centering
    \begin{tabular}{lr@{~~}r@{~~}r@{~~}r}
                     & \llap{Correct} & Clash & Non. & Oth. \\
\hline
Reference            &    16  &                1  &  -        &  -     \\
\onlB{}              &     9  &                8  &  -        &  -     \\
\docTr{}             &     8  &                7  &        2  &  -     \\
\onlY{}              &     8  &                7  &  -        &      2 \\
\TrEighteen{}        &     8  &                7  &        1  &      1 \\
\TrMarian{}          &     8  &                6  &        1  &      2 \\
\tartu{}             &     8  &                6  &        2  &      1 \\
\onlA{}              &     7  &                8  &  -        &      2 \\
\onlX{}              &     7  &                8  &  -        &      2 \\
\TrNineteen{}        &     7  &                7  &  -        &      3 \\
\uedin{}             &     7  &                5  &        1  &      4 \\
\onlG{}              &     6  &                7  &        1  &      3 \\
    \end{tabular}
    \caption{How the systems were referring to the contracting parties. ``Correct'' indicates an appropriate and consistent translation. ``Clash'' indicates that the translation wrongly refers to the other party. ``Non.'' are cases when the original English word appeared in MT output and ``Oth.'' are other translations; these are also confusing because the identity with the correct party is not maintained.}
    \label{tab:reference:correctness}
\end{table}

Our analysis so far does not sufficiently highlight the most severe flaw of all the MT systems. The problem concerns a clear way of referring to the contracting parties, i.e. the translation of the terms ``tenant'' and ``lessee''. All the systems translated almost all occurrences of these terms using one word only, ``nájemce'', which causes a lot of confusion to any reader (including native Czech speakers). The problem which occurred here arose from the fact that there are actually three common roles and two types of agreements in apartment renting. Commonly, the contracting parties are:
\begin{itemize}
    \item landlord---tenant = pronajímatel---nájemce in the case when the landlord is the owner of the property;
    \item tenant---lessee = nájemce---podnájemce for the sublease agreement, i.e. when the owner is not directly involved in the agreement.
\end{itemize}
The common translation in training corpora or dictionaries of the term ``lessee'' is apparently ``nájemce'' which is possible, but only if the term ``tenant'' is not used in the document as well. Should this happen, ``lessee'' needs to be translated as ``podnájemce'' to avoid confusion.

\cref{tab:reference:correctness} details the performance of the systems in this respect. Each line sums up to 17 mentions of either of the two contracting parties.
We see that the reference translation made only one error by using the wrong term while all the other systems cause a term clash (using the same term for both parties) in half of the cases. This, in fact, corresponds to all the mentions of the second party and  \emph{all these translations by all the systems} are thus completely wrong.

\section{Test Suite Availability}
\label{availability}

SAO Test Suite is available under CC-BY-SA at:
\begin{center}
\url{https://github.com/ELITR/wmt19-elitr-testsuite}
\end{center}

\section{Conclusion}
\label{conclusion}


We presented a test suite of Czech, English, German, Polish and Slovak documents
from the auditing domain and used its English-Czech-German tri-parallel part in
the WMT19 Translation Shared Task. We also added one more document type, namely a sublease agreement.

Despite the fact that the participating MT systems were trained for a rather
general domain of news articles, many of them perform very well \emph{on general terms}. Our detailed
manual evaluation used criteria similar to those used in the scoring of GCSE essays of the Czech language.

An important observation in our study was that a thorough domain knowledge is
necessary to assess the correctness of the translation, esp. in terms of lexical
choices, and that the reference translations are insufficient for the task. Our
impression is that automatic MT evaluation is effectively useless for assessing
terminological subtleties, esp. with one reference translation only.
We find this observation particularly important for future research
directions, because none of the MT systems are trained in a way which could
directly address such subtle issues.
Terminology lists may be a good help for both MT and MT evaluation but we anticipate that the only
practically
possible ultimate
solution for translation would be an interactive system supporting a domain expert in manual correction of terminological choices.

As for the translations of the Sublease Agreement, even though the dispersion in
the number of errors is huge---varying from 21 errors
(CUNI-Transformer-T2T-2018) to 125 errors online-X---the number of errors alone
is not as indicative of the practical usability of the translation. The main
problem was that \emph{all} the systems made the same (and from the readers'
perspective, the most severe) translation error by translating the terms ``tenant'' and ``lessee'' using the same Czech word ``nájemce'', which made the whole text incomprehensible. Other observed mistakes needed rather cosmetic adjustments, except for the occasions where the system forgot a whole sentence or the rest of a paragraph.

We released the texts of the test suite for future use and we are also happy to share our annotation protocols, but as of now, we cannot provide any novel automatic evaluation of MT on this test suite.

\section*{Acknowledgments}

This study was supported in parts by the grants H2020-ICT-2018-2-825460 (ELITR) and H2020-ICT-2018-2-825303 (Bergamot) of the European Union and Czech Science Foundation (grant n.\ 19-26934X, NEUREM3).



We are very grateful to our colleagues and students Ivana Kvapilíková,
Ján Faryad,
Lukáš Kyjánek, and
Simon Will for their help with the revision of the German alignment.

\bibliography{biblio}
\bibliographystyle{acl_natbib}

\end{document}